\documentclass{llncs}
\usepackage{mathrsfs}
\usepackage{amsfonts}
\usepackage{xcolor}
\usepackage[pdftex]{graphicx}
\usepackage{caption}
\usepackage{amssymb}
\usepackage{mathrsfs}
\usepackage{amsfonts}

\usepackage{multirow}
\usepackage{booktabs}
\usepackage{amsmath}
\usepackage{caption}
\usepackage{makecell}
\usepackage{subfigure}
\usepackage{mathrsfs}
\usepackage{color}
\usepackage{amsmath}
\usepackage{amssymb}

\usepackage{cite}
\usepackage{hyperref}
\begin{document}

\mainmatter  % start of an individual contribution
\author{Huaying Hao\inst{1}
\and Huazhu Fu\inst{2*}
\and Yanwu Xu\inst{3}
\and Jianlong Yang\inst{1}
\and Fei Li\inst{5}
\and Xiulan Zhang\inst{5}
\and Jiang Liu\inst{4}
\and Yitian Zhao\inst{1*}
}

\institute{Cixi Institute of Biomedical Engineering, Ningbo Institute of Materials Technology and Engineering, Chinese Academy of Sciences, Ningbo, China, \email{yitian.zhao@nimte.ac.cn}
\and Inception Institute of Artificial Intelligence, \email{hzfu@ieee.org}
\and Baidu Inc
\and Southern University of Science and Technology
\and State Key Laboratory of Ophthalmology, Zhongshan Ophthalmic Center, Sun Yat-sen University, Guangzhou
}
% first the title is needed
\title{Open-Narrow-Synechiae Anterior Chamber Angle Classification in AS-OCT Sequences}

\maketitle

\begin{abstract}
Anterior chamber angle (ACA) classification is a key step in the diagnosis of angle-closure glaucoma in Anterior Segment Optical Coherence Tomography (AS-OCT). Existing automated analysis methods focus on a binary classification system (i.e., open angle or angle-closure) in a 2D AS-OCT slice. However, clinical diagnosis requires a more discriminating ACA three-class system  (i.e., open, narrow, or synechiae angles) for the benefit of clinicians who seek better to understand the progression of the spectrum of angle-closure glaucoma types.
To address this, we propose a novel sequence multi-scale aggregation deep network (SMA-Net) for open-narrow-synechiae ACA classification based on an AS-OCT sequence. In our method, a Multi-Scale Discriminative Aggregation (MSDA) block is utilized to learn the multi-scale representations at slice level, while a ConvLSTM is introduced to study the temporal dynamics of these representations at sequence level. Finally, a multi-level loss function is used to combine the slice-based and sequence-based losses. The proposed method is evaluated across two AS-OCT datasets. The experimental results show that the proposed method outperforms existing state-of-the-art methods in applicability, effectiveness, and accuracy. We believe this work to be the first attempt to classify ACAs into open, narrow, or synechiae types grading using  AS-OCT sequences.

%To address this,in this work, we present a new approach that is not only able to grade the ACAs into open and angle-closure which follows the conventional approaches, but also is capable of the classification of ACAs into a more precise manner: open, narrow, and synechia, so as to benefit clinicians who seek better to understand the progression of the spectrum of angle-closure glaucoma types.
% We propose a novel sequence multi-scale aggregation deep network (MA-Net) for learn multi-scale discriminative representation in AS-OCT images. A new multi-loss function ($SV-loss$), which includes 2D slice loss and 3D volume loss, is introduced to study the temporal dynamics of these representations from image sequences for grading of ACAs into open, narrow, and synechia. The proposed method is evaluated across two AS-OCT datasets and the experimental results show that the proposed method outperforms existing state-of-the-art methods in applicability, effectiveness, and accuracy.
\end{abstract}

\section{Introduction}

\begin{figure}[t]
\centering
\includegraphics[width = 1\linewidth]{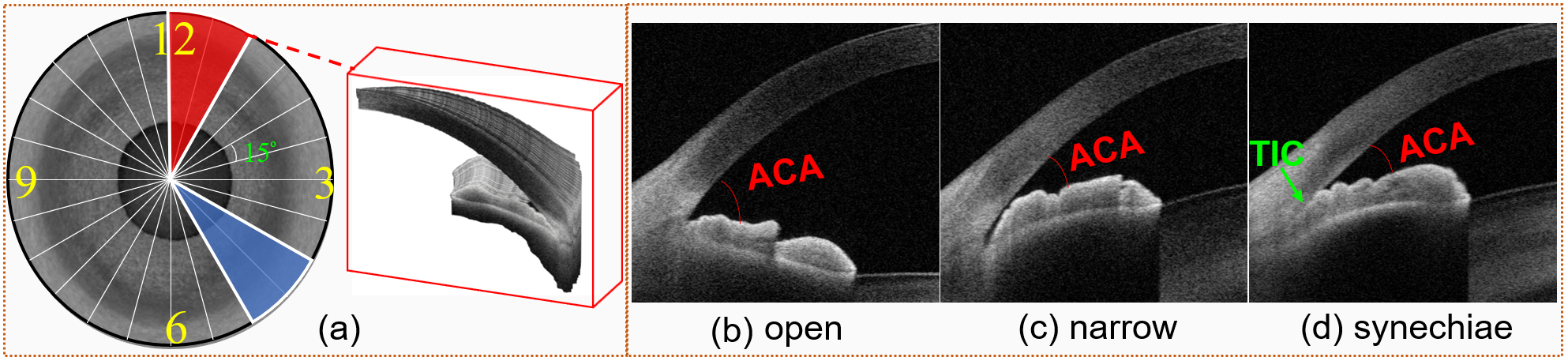}
\caption{(a) Visual demonstration of manual annotations by gonioscopy. ACA angles in the 12-1 o'clock region (red) viewed with synechiae, and 4-5 o'clock (blue) viewed with narrow ACAs, and a sequence of AS-OCT images in a random $15^{\circ}$ radiant area. (b)-(d)  Image samples of open-narrow-synechiae ACA grading.}
\label{fig.1}
\vspace{-8pt}
\end{figure}

Glaucoma is one of the most significant causes of irreversible blinding  worldwide, and primary angle-closure glaucoma (PACG) is the major cause of blindness in Asian populations~\cite{tham2014global}.  Anterior Segment Optical Coherence Tomography (AS-OCT) imaging, as a non-contact and non-invasive tool, is widely used to observe cross-sections of the anterior segment structure, to assist ophthalmologists in early screening and accurate assessment of PACG~\cite{fu2019deep}, as shown in Fig.~\ref{fig.1}(a).
%In particular, as a non-contact and non-invasive tool, AS-OCT has been shown to provide a clear cross-sections of anterior segment structures, as shown in Fig. \ref{fig.1} (a).
% Ginoscopy and Anterior Segment Optical Coherence Tomography (AS-OCT) are two common imaging techniques to observe the anterior chamber angle (ACA), so as to further assist ophthalmologists to early screening and accurate assessment. In particular, as a non-contact and non-invasive tool, AS-OCT has been shown to provide a clear cross-sections of anterior segment structures, as shown in Fig. \ref{fig.1} (a).

According to the configuration of the anterior chamber angle (ACA), cases of glaucoma may first be classified into two types: primary open angle glaucoma (Fig.~\ref{fig.1}(b));  and angle-closure glaucoma  (Fig.~\ref{fig.1}(c-d)). Drawing on epidemiological research~\cite{foster2002definition}, clinicians have suggested that angle-closure glaucoma may be further divided into different sub-stages: primary angle closure suspect (PACS), and primary angle-closure/primary angle-closure glaucoma (PAC/PACG). PACS denotes the eye with \textbf{narrow} but non-adhesive ACA, as shown in Fig.~\ref{fig.1}(c), while PAC/PACG denotes  the eye with an occludable ACA, \emph{a.k.a}, \textbf{synechiae}, which leads to the presence of trabecular iris contact (TIC)~\cite{fu2019angle}, as shown in Fig.~\ref{fig.1}(d). This secondary level of grading would benefit clinicians in better  understanding the progression of the spectrum of angle-closure glaucoma types. Moreover, surgical treatment during PACS can open any non-firm adhesion between the peripheral iris and the trabecular meshwork, which may avoid or alleviate the permanent adhesion seen in the progression to PACG~\cite{shang2019automated}. 
Therefore, the accurate identification of open-narrow-synechiae ACAs is potentially important in guiding clinical management at different stages of angle-closure glaucoma.

Several methods have been proposed to enable automatic classification of open angle and angle-closure ACAs from AS-OCT images.
%, which can be divided into two main categories: clinical measurement-based methods and feature-based methods. Clinical measurement-based methods first segment the AS-OCT image, and then identify the subtypes of glaucoma based on clinical parameters. 
 Xu \emph{et al}.~\cite{xu2012anterior,xu2013automated} began by localizing the ACA region, and then classified the ACA into open angle or angle-closure based on visual features present in their AS-OCT images. Fu \emph{et al}.~\cite{fu2017segmentation} proposed a data-driven approach to integrate AS-OCT segmentation, clinical parameter measurements, and glaucoma screening. However, clinical measurement-based methods rely heavily on precise segmentation of the AS-OCT structure. %Feature-based methods, on the other hand, first extract ACA region, and then classify glaucoma subtypes based the representations of region. 
Recently, deep learning-based methods have demonstrated superior performance in ACA classification~\cite{fu2019deep,xu2019deep}. Fu \emph{et al}.~\cite{fu2019angle}  proposed a multi-context deep network, in which parallel convolutional neural networks are applied to ACA regions and at corresponding scales known to be informative for clinically diagnosing angle-closure glaucoma.  Xu \emph{et al}.~\cite{xu2019deep} employed deep learning classifiers for automated detection of gonioscopic angle closure and primary angle closure.
%Fu \emph{et al}.~\cite{fu2019angle} further introduced a Multi-Level Deep Network, which includes multiple parallel sub-networks to learn multi-level representations from the multiple regions known to be informative for angle-closure detection in an AS-OCT image. 
%In these deep feature-based methods, the final classification results are determined by model integration from global and local regions to improve the ability of multi-scale learning.
However, multi-level network learning is costly in terms of storage capacity, and slow to proceed to inference. In addition,  it depends on a proper integration model, in the absence of which it is less effective than a single model. 
%To address the above issues, in this paper,  we present a novel building block to enhance multi-scale representation ability at the convolutional neural network (CNN) layer level, which can be easily integrated with CNN modules in AS-OCT images for glaucoma classification.
%Xu \emph{et al}.~\cite{xu2019deep} employed deep learning classifiers for automated detection of gonioscopic angle closure and primary angle closure disease(PACD). 

%This is because it is easy to distinguish open angle  from closed angle by AS-OCT images, however, it is difficult to classify narrow angle area and synechiae angle area based solely on a single static AS-OCT image. The reason is  that a synechiae angle area may appear dynamic adhesion, and classifier based on static AS-OCT images may not predict accurate result.

%Inspired by the processing of the gonioscopy examination, we address three subtypes of glaucoma classification as image sequences classification and propose a new architecture, which includes CNNs with the proposed block and Convolutional Long short-term memory(ConvLSTM) neural network.

%The proposed novel deep network is able to classify the ACAs into both two (open/angle-closure) and three (open/narrow/synechiae) subtypes. 
All the aforementioned methods focus on the binary classification of open angle and angle-closure in a 2D slice. The open-narrow-synechiae ACA classification based on an AS-OCT sequence has been rarely explored, despite its significance in  understanding disease progression~\cite{hao2019anterior}. 
Inspired by the procedures of the dynamic gonioscopy examination, in which ophthalmologists move the gonioscope counterclockwise and make  an annotation every $15^{\circ}$ (see left row of Fig. \ref{fig.1}(a)), we address the open-narrow-synechiae ACA classification as an image sequence classification problem. The main contributions of this paper are summarized as follows.
\textbf{(1)} We develop a  sequence multi-scale aggregation deep network (SMA-Net) for discriminating the temporal dynamics of features in order to classify ACAs into the open-narrow-synechiae grading, using  AS-OCT sequences.
To best of our knowledge, this is the first work in the area of automated open-narrow-synechiae ACA classification. 
\textbf{(2)} A multi-scale discriminative aggregation (MSDA) block is designed to extract the multi-scale representations at slice level, and a ConvLSTM is employed to study the temporal dynamics of these representations at sequence level. 
\textbf{(3)} A new multi-loss function is used to combine slice-based and sequence-based losses, so as to  extract  spatial and temporal features from AS-OCT sequences.
\textbf{(4)} Our proposed method outperforms existing state-of-the-art methods in applicability, effectiveness, and accuracy on two AS-OCT datasets (one public and one private dataset).

\section{Methodology}

\begin{figure}[!t]
\centering{
\includegraphics[width=12cm]{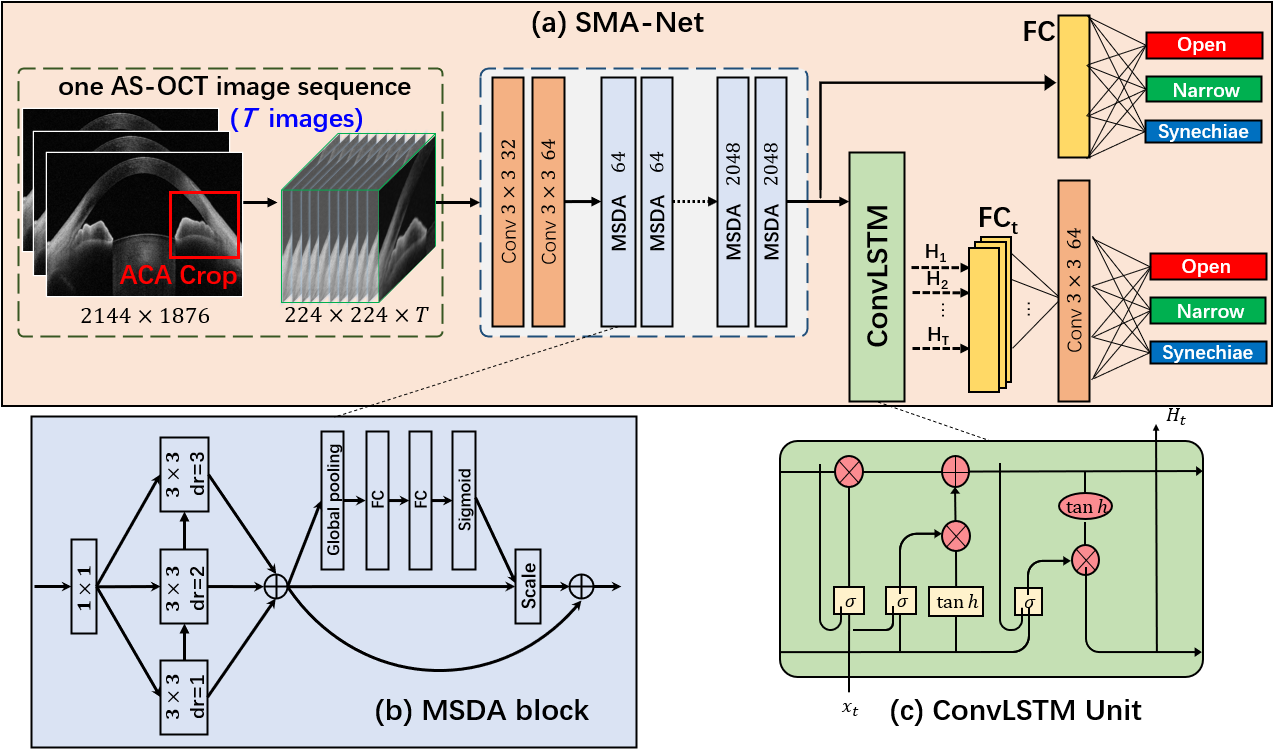}
}
\caption{Overview of our SMA-Net for open-narrow-synechiae ACA classification based on an AS-OCT sequence.}
\label{fig.2}
\end{figure}

Fig.~\ref{fig.2}(a) illustrates the framework of our SMA-Net. Given an AS-OCT sequence, first, a coarse-to-fine method~\cite{hao2019anterior} is utilized to localize the ACA regions with sizes of $448\times 448$ pixels for each AS-OCT slice, which is the most useful discriminative area for glaucoma classification. 
Then, a cropped ACA sequence with of size $224\times 224\times T$ is obtained and fed to MSDA blocks to extract a sequence of feature maps at slice level. Here $T$ denotes the scan number of the AS-OCT within a $15^{\circ}$ radiant area. 
Finally, a stacked 2-layer ConvLSTM is employed to study the temporal dynamics of these representations at sequence level, which produces the prediction of  open-narrow-synechiae ACA. 

\subsection{Multi-Scale Discriminative Aggregation Block}

In our method, each MSDA block consists of a depthwise separable convolutional module and an aggregation gate module. The structure of a depthwise separable convolutional module includes one $1\times1$ convolutional layer and three $3\times3$ atrous convolutional layers, as shown in Fig.~\ref{fig.2}(b). The separable convolution not only reduces the number of parameters, but improves feature learning ability of the network~\cite{chollet2017xception,yu2018deep}.  
$\textbf{x}$ denotes the  output of the $1\times1$ convolution, and $\mathbf{f}_{i}$ corresponding to $3\times3$ atrous convolutional layers are used to enlarge the receptive field,  with different dilation rates $i\in \{ 1, 2, 3\}$. Moreover, Batch Normalization and ReLU activation are used for each convolutional layer. The feature $\textbf{x}$ is added to the output of $\mathbf{f}_{i-1}$, and then fed into $\mathbf{f}_{i}$, as 
\begin{equation}\small
\begin{aligned} 
\mathbf{y}_{i}=\left\{\begin{array}{ll}{\mathbf{f}_{i}\left(\mathbf{x}\right)},& {i=1} \\ {\mathbf{f}_{i}\left(\mathbf{x}+\mathbf{y}_{i-1}\right)}, & {i>1}\end{array}\right. ,
\end{aligned}
\end{equation}
where $\mathbf{y}_{i}$ denotes the output of $\mathbf{f}_{i}$. Finally, multi-scale convolutional features are aggregated as the output of the depthwise separable convolutional module. Compared with existing multi-scale blocks, our block effectively reduces the number of parameters within the hierarchical aggregation structure.

In order to combine different scale representations  more effectively, an aggregation gate is also introduced, using a Squeeze and Excitation (SE) module to obtain discriminative representation. With the output of the depthwise separable convolutional module, the SE module is used in our block followed the coarse fusion to the new output. More specifically, the SE module is composed of a global average pooling layer and a Multi-Layer Perceptron (MLP) with a ReLU-activated hidden layer, followed by the sigmoid activation.  Note that the SE module follows a self-attention architecture, which can selectively enhance discriminative representation. In contrast to single feature addition fusion, this more discriminative fusion can achieve dynamic weighted distribution in the channels of multi-scale branches.

%\subsection{Classification of open angle and angle-closure}
%Our angle-closure glaucoma classification system includes two stages: ACA region localization and MA-Net, as shown in the first stream of Fig. \ref{fig.2}. First, a coarse-to-fine method~\cite{hao2019anterior} is utilized to localize the ACA region with sizes of $448\times 448$ from an AS-OCT image, which is the most discriminative area for glaucoma classification. Then, the ACA regions are resized to $224\times 224$ and input MA-Net for identification of open angle and angle-closure glaucoma. Moreover, in our MA-Net, as shown as in Fig. \ref{fig.2}, we design our network based on Xception architecture~\cite{chollet2017xception} and replace the $3\times 3$ convolution with our MSDA block, which can learn multi-scale representation and integrate the discriminative representation with the output. The output maps of our MA-Net are input a fully connected layer for obtaining final classification results. Finally, standard binary cross entropy is employed for training our network

\subsection{Open-narrow-synechiae ACA classification}

In our method, Xception architecture~\cite{chollet2017xception} is used as the backbone. We replace the $3\times 3$ convolution with our MSDA block, which can learn multi-scale representation and integrate the discriminative representation with the output. With the slice-based features from MSDA blocks, a stacked 2-layer ConvLSTM, with 1024 hidden units in each of the cells, is used to process feature maps and generate a sequence of ConvLSTM states $\left \{ \mathbf{H}_{1},\mathbf{H}_{2},\cdots,\mathbf{H}_{T} \right \}$, which can learn spatial and temporal information by spatial dependencies, as shown in Fig.~\ref{fig.2}(c). At each position $t$, the output state $\mathbf{H}_{t}$ is fed into a global average pooling and a fully connected (FC) layer that computes the estimated probabilities $\left \{ \mathbf{Y}_{1},\mathbf{Y}_{2},\cdots,\mathbf{Y}_{T} \right \}$. 

%In this section, our open-narrow-synechiae angle classification system includes three stages, ACA region localization, MA-Net and ConvLSTM, as shown in the second of Fig. \ref{fig.2}. First, a coarse-to-fine method~\cite{hao2019anterior} is utilized to localize the ACA regions with sizes of $448\times 448$ from AS-OCT images of an image sequence, which is the most discriminative area for glaucoma classification. 
%Then, an ACA image sequence with $224\times 224\times11$ is obtained and input MA-Net to extract a sequence of feature maps$\left \{ \mathbf{X}_{1},\mathbf{X}_{2},\cdots,\mathbf{X}_{t} \right \}$, where $t=11$. In our MA-Net, as shown as in Fig. \ref{fig.2}, we design our network based on Xception architecture~\cite{chollet2017xception} and replace the $3\times 3$ convolution with our MSDA block, which can learn multi-scale representation and integrate the discriminative representation with the output. Finally, we use a stacked 2-layer ConvLSTM, which has 1024 hidden units in each of cells, to process feature maps from MA-Net in a sequential way and generate a sequence of ConvLSTM states $\left \{ \mathbf{H}_{1},\mathbf{H}_{2},\cdots,\mathbf{H}_{t} \right \}$, which can learn spatial and temporal information by spatial dependencies. 

In this work, we consider a multi-level loss function  to combine slice-based and sequence-based losses. To optimize the SMA-Net and avoid gradient disappearance, we propose to use cross entropy loss as our slice-based loss. To be specific, we use global average pooling and an FC layer to generate estimated probabilities $\mathbf{Y}^{s}_{t}$ from each slice $t$. To this end, the slice-based loss is defined as:
\begin{equation}\small
\mathcal{L}_{\mathrm{2D}}= \sum_{t=1}^{T}\left \{ \mathcal{L}_{CE}\left ( \mathbf{Y}^{s}_{t},\mathbf{Y}^{*} \right ) \right \},
\end{equation}
where $\mathcal{L}_{CE}$ represents cross entropy loss, and $\mathbf{Y}^{*}$ denotes the ground truth of the input AS-OCT sequence. Note that in our method, all AS-OCT slices have the same labels within the whole sequence.

To improve performance, we use a weighted ensemble (WE) method to optimize the final result by these estimated probabilities. We first concatenate a sequence of estimated probabilities to obtain a new descriptor with dimension $T\times3$. We then perform a 1-D convolution operation with on the new descriptor, and feed the result to the FC layer for final classification $\mathbf{Y}_{f}$, which can weight predictions of $T$ state. Finally, we define a new loss-based cross entropy as our sequence-based loss function:
\begin{equation}\small
\mathcal{L}_{\mathrm{3D}}= \sum_{i=1}^{t}\left \{ \mathcal{L}_{CE}\left ( \mathbf{Y}_{i},\mathbf{Y}^{*} \right ) \right \}.
\end{equation}
Therefore, the overall loss function of our network is as: 
\begin{equation}\small
\mathcal{L}_{\mathrm{SV}}= \mathcal{L}_{\mathrm{2D}}+ \lambda \mathcal{L}_{\mathrm{3D}},
\end{equation}
where $\lambda$ is the balance weight ($\lambda = 1$ in our experiment).

\section{Experimental Results}

The proposed architecture was implemented using the publicly available Pytorch Library. In the training phase, we employed an Adam optimizer to optimize the deep model. We used a gradually decreasing learning rate, starting from $0.0001$. In addition, online data enhancement was employed to enlarge the training data, which includes brightness, color, contrast and sharpness transformations, and we set a random seed from 1 to 4 for enhancement. 

\subsection{Performance of open-narrow-synechiae ACA classification}

%By means of AS-OCT imaging technology, such as provided by the CASIA-2 AS-OCT machine (Tomey Inc., Japan), which captures eleven continuous scans in a $15^{\circ}$ radiant area,  we can analyze this portion of the complete 3D volume as a sequence of 2D images.

\textbf{{Private dataset:} }
%A local AS-OCT dataset was used in this work to valid the proposed method in grading ACAs to open-narrow-synechiae angle. 
All AS-OCT volumes in our dataset were captured  by a CASIA-2 machine (Tomey Inc., Japan)  from 66 eyes with open-narrow-synechiae grading (human expert annotations of images derived from by dynamic gonioscopic examinations).
%, and the structure of the anterior chamber appears at a relatively consistent position among AS-OCT images in practice.  
Each volume contains 128 AS-OCT images, and each slice in a volume is $2144\times1876$ pixels. 
%It is very worth noting this clinical dataset has two sub-sets, referred as \textbf{Dark} and \textbf{Light} sets in this paper. Imaging of each patients was performed first in dark conditions and then under bright conditions using a measured standardized light source: dark (0.4 lux) and light (104 lux) illumination conditions. Pupil has different responses under various illumination conditions, and the changing of the pupil size may stretch the iris and lead to the morphological changes of ACA. Fig \ref{fig.7} demonstrates two ACAs under different illumination conditions. It may be seem clearly that the narrowed angle has different appearances when the pupil size changed, while synechia keeps relative consistent appearance.  We tend to use pupil changing to simulate the pressure by the goniolens, which can push the angle open and help determine the true angle configuration.  
A senior ophthalmologist annotated  each $15^{\circ}$ segment of the  ACAs of an eye, yielding 24 annotations for each single eye, and resulting in a total of 1584 annotations for this dataset. In light of this, we formed the AS-OCT slices ($T=11$ slices) in each $15^{\circ}$ region as one sequence. Finally, a total of 1584 image sequences are generated, in which 504 sequences are annotated with open ACA, 742 sequences are with narrow ACA, and 338 sequences contain synechiae ACA. In our experiments, the dataset was equally and randomly divided into training and testing sets, taking into account the integrity of data from each eye, so that two ACAs from the same eye would not be separated between the training and testing sets.

\begin{table}[t]
\centering
\caption{Classification of the open-narrow-synechiae ACAs by different methods on the private dataset.}
\setlength{\tabcolsep}{3.5mm}{
\begin{tabular}{lllllll} 
\hline
\multicolumn{1}{l||}{Methods} & \multicolumn{1}{c|}{\textit{kappa}} & \multicolumn{1}{c|}{\textit{F1}} & \multicolumn{1}{c|}{\textit{B-Acc}} & \multicolumn{1}{c|}{\textit{Sen}}& \multicolumn{1}{c}{\textit{Spe}}\\
\hline 
\multicolumn{1}{l||}{ResNet-34~\cite{he2016deep}} & \multicolumn{1}{c|}{0.6766} & \multicolumn{1}{c|}{0.7527}& \multicolumn{1}{c|}{0.8188} & \multicolumn{1}{c|}{0.7485}& \multicolumn{1}{c}{0.8891}\\
\multicolumn{1}{l||}{MSCNN~\cite{hao2019anterior}} & \multicolumn{1}{c|}{0.6773} & \multicolumn{1}{c|}{0.7498} & \multicolumn{1}{c|}{0.8171} & \multicolumn{1}{c|}{0.7455}& \multicolumn{1}{c}{0.8887}\\
\multicolumn{1}{l||}{Xception~\cite{chollet2017xception}} & \multicolumn{1}{c|}{0.7252} & \multicolumn{1}{c|}{0.7835}& \multicolumn{1}{c|}{0.8393} & \multicolumn{1}{c|}{0.7752}& \multicolumn{1}{c}{0.9035}\\
\multicolumn{1}{l||}{MA-Net} & \multicolumn{1}{c|}{0.7477} & \multicolumn{1}{c|}{0.8121} & \multicolumn{1}{c|}{0.8600} & \multicolumn{1}{c|}{0.8063}& \multicolumn{1}{c}{0.9137}\\
\hline
\multicolumn{1}{l||}{C3D~\cite{tran2015learning}} & \multicolumn{1}{c|}{0.7489} & \multicolumn{1}{c|}{0.8115} & \multicolumn{1}{c|}{0.8532} & \multicolumn{1}{c|}{0.8048}& \multicolumn{1}{c}{0.9136}\\
\multicolumn{1}{l||}{I3D~\cite{carreira2017quo}} & \multicolumn{1}{c|}{ 0.7662} & \multicolumn{1}{c|}{0.8171} & \multicolumn{1}{c|}{0.8619} & \multicolumn{1}{c|}{0.8073}& \multicolumn{1}{c}{0.9166}\\
\multicolumn{1}{l||}{S3D~\cite{xie2018rethinking}} & \multicolumn{1}{c|}{0.7431} & \multicolumn{1}{c|}{0.8007} & \multicolumn{1}{c|}{0.8567} & \multicolumn{1}{c|}{0.8016}& \multicolumn{1}{c}{0.9119}\\
\hline
%\multicolumn{1}{l||}{ResNet-34+ConvLSTM} & \multicolumn{1}{c|}{0.7689} & \multicolumn{1}{c|}{0.8281} & \multicolumn{1}{c|}{0.8714}& \multicolumn{1}{c|}{0.8220}& \multicolumn{1}{c}{0.9208}\\
\multicolumn{1}{l||}{Our SMA-Net} & \multicolumn{1}{c|}{\textbf{0.7931}} & \multicolumn{1}{c|}{\textbf{0.8459}} & \multicolumn{1}{c|}{\textbf{0.8829}}& \multicolumn{1}{c|}{\textbf{0.8371}}& \multicolumn{1}{c}{\textbf{0.9282}}\\
\hline
\end{tabular}}
\label{table1}
\end{table}

In order to demonstrate the superiority of the proposed method for classification of open-narrow-synechiae angle, we carry out a comprehensive comparison between the proposed method and the following state-of-the-art methods:
(1) 2D deep models: Multi-Scale Regions Convolutional Neural Networks (MSCNN)~\cite{hao2019anterior}; Resnet-34~\cite{he2016deep}; Xception~\cite{chollet2017xception}; and MA-Net (Xception with our MSDA block).  3D deep models: C3D~\cite{tran2015learning}; I3D~\cite{carreira2017quo}; and S3D~\cite{xie2018rethinking}. Following the standard performance assessment protocol for multi-class classification\cite{Annunziata16}, we use weighted sensitivity (\textit{Sen}), specificity (\textit{Spe}), and balanced accuracy (\textit{B-acc}). In order to reflect the trade-offs between \textit{Sen} and \textit{Spe}  and evaluate the quality of our classification results, the $\textit{kappa}$ analysis~\cite{kappa} and \textit{F1}~\cite{F1} score were also  provided.
%In our experiment, Table \ref{table1} and Table \ref{table2} are used to evaluate the performance of all methods in open-narrow-synechiae angle classification. 
Table \ref{table1} reports the classification performance of different methods. It may be observed that our SMA-Net yields best performance in terms of all metrics when compared to either 2D or 3D deep learning-based methods. The probable reason for this is that the proposed networks can learn rich and discriminative representation from both local features (2D image features) and global geometry (image sequence).

\begin{figure*}[t]
\centering{
\includegraphics[width=12.5cm]{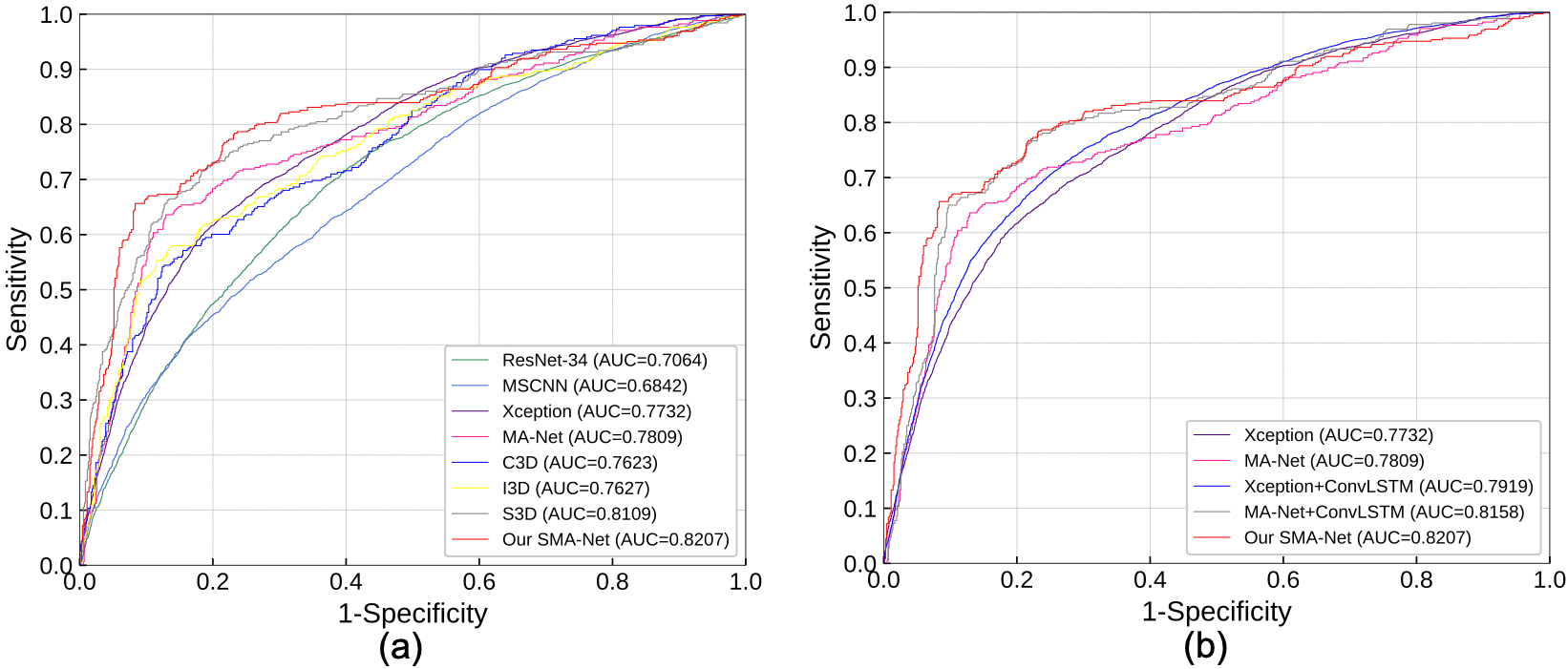}
}
\caption{AUC curves of different approaches in classifying narrow, and synechiae ACA. (a) Comparison of state-of-the-art 2D and 3D deep networks. (b) Ablation study of our model. }
\label{fig.3}
\vspace{-12pt}
\end{figure*}

Because binary open angle and angle-closure classification is a relatively straightforward and easy task, our method and other networks achieve similar high performances (AUC=0.998). Therefore,  only the AUC curves in distinguishing narrow and synechiae angles are illustrated in Fig. \ref{fig.3}(a). As expected, our method still produced the best performance on the classification of narrow and  synechiae angles, with AUC=0.8207. Overall, all 3D networks achieved relatively higher performances than 2D networks, since a 3D deep network can learn spatial representation from an image sequence. 

%
%It may be observed from Table \ref{table1} and  Fig. \ref{fig.3}(a) that our MA-Net yields better classification performance than other 2D deep learning-based method. In addition, 3D deep network obtained higher performance than 2D network. This is because 3D deep network can learn spatial representation from image sequence. Overall, our method outperforms the other networks in terms of all metrics and with kappa of 0.7922 and auc of 0.8207 in the Local dataset. A possible reason for this is that our method can learn rich local features (image features) and global features (spatial features) from image sequence. And compared with 3D network, when the image sequence length is 11, our method may not cause  disappearance of spatial feature. 

To evaluate the effectiveness of each module in our network, we provide an ablation study, and the results are reported in Table \ref{table2} and Fig. \ref{fig.3}(b). Results show that the Xception+ConvLSTM and MA-Net+ConvLSTM method clearly outperformed the and Xception and MA-Net alone, with  improvement of about 4.53\% and 3.43\% in \textit{kappa}, and about 1.87\% and 3.49\% in AUC. This demonstrates that the ConvLSTM learns discriminative representations from an AS-OCT image sequence, and is capable of preserving the spatial information, so memorizing the change in appearance that corresponds to motion information (neighboring slices), thereby improving performance in separating narrow angle and fully closed angle.
Table \ref{table2} and Fig. \ref{fig.3}(b) also reveal that after $\mathcal{L}_{\mathrm{SV}}$ was applied, classification performance of our SMA-Net (MA-Net+ConvLSTM+$\mathcal{L}_{\mathrm{SV}}$)  improved significantly, with an improvement of approximately 4.53\% in \textit{kappa} and 0.49\% in AUC, respectively. This is because with $\mathcal{L}_{\mathrm{SV}}$, the ConvLSTM explores the temporal dynamics of appearance features of an AS-OCT sequence, and these features are further aggregated for classification purposes.

\begin{table}[t]
\centering
\caption{Classification performance of the open-narrow-synechiae ACAs by different module combinations on private dataset.}
\setlength{\tabcolsep}{2mm}{
\begin{tabular}{lllllll} 
\hline
\multicolumn{1}{l||}{Methods} & \multicolumn{1}{c|}{$\textit{kappa}$} & \multicolumn{1}{c|}{\textit{F1}} & \multicolumn{1}{c|}{\textit{B-acc}} & \multicolumn{1}{c|}{\textit{Sen}}& \multicolumn{1}{c}{\textit{Spe}}\\
\hline 
\multicolumn{1}{l||}{Xception~\cite{chollet2017xception}} & \multicolumn{1}{c|}{0.7252} & \multicolumn{1}{c|}{0.7835}& \multicolumn{1}{c|}{0.8393} & \multicolumn{1}{c|}{0.7752}& \multicolumn{1}{c}{0.9035}\\
\multicolumn{1}{l||}{MA-Net} & \multicolumn{1}{c|}{0.7477} & \multicolumn{1}{c|}{0.8121} & \multicolumn{1}{c|}{0.8600} & \multicolumn{1}{c|}{0.8063}& \multicolumn{1}{c}{0.9137}\\
\hline
%\multicolumn{1}{l||}{ResNet-34+ConvLSTM} & \multicolumn{1}{c|}{0.7689} & \multicolumn{1}{c|}{0.8281} & \multicolumn{1}{c|}{0.8714}& \multicolumn{1}{c|}{0.8220}& \multicolumn{1}{c}{0.9208}\\
\multicolumn{1}{l||}{Xception+ConvLSTM} & \multicolumn{1}{c|}{0.7705} & \multicolumn{1}{c|}{0.8238} & \multicolumn{1}{c|}{0.8664}& \multicolumn{1}{c|}{0.8135}& \multicolumn{1}{c}{0.9194}\\
\multicolumn{1}{l||}{MA-Net+ConvLSTM} & \multicolumn{1}{c|}{0.7820} & \multicolumn{1}{c|}{0.8377} & \multicolumn{1}{c|}{0.8780}& \multicolumn{1}{c|}{0.8310}& \multicolumn{1}{c}{0.9250}\\
\multicolumn{1}{l||}{MA-Net+ConvLSTM+$\mathcal{L}_{\mathrm{SV}}$} & \multicolumn{1}{c|}{\textbf{0.7931}} & \multicolumn{1}{c|}{\textbf{0.8459}} & \multicolumn{1}{c|}{\textbf{0.8826}}& \multicolumn{1}{c|}{\textbf{0.8371}}& \multicolumn{1}{c}{\textbf{0.9282}}\\
\hline
\end{tabular}}
\label{table2}
\vspace{-12pt}
\end{table}

\begin{table}[t]
\centering
\caption{Performance of compared methods on AGE dataset.}
\setlength{\tabcolsep}{3mm}{
\begin{tabular}{lllllll}  
\hline
\multicolumn{1}{l||}{Methods} & \multicolumn{1}{c|}{AUC}& \multicolumn{1}{c|}{\textit{Sen}}& \multicolumn{1}{c}{\textit{Spe}}\\
\hline 
\multicolumn{1}{l||}{Resnet-34~\cite{he2016deep}} & \multicolumn{1}{c|}{0.990959} & \multicolumn{1}{c|}{0.859375} & \multicolumn{1}{c}{0.999219}\\
\multicolumn{1}{l||}{Inception-V3~\cite{xia2017inception}} & \multicolumn{1}{c|}{ 0.999285} & \multicolumn{1}{c|}{0.918750} & \multicolumn{1}{c}{0.998438}\\
\multicolumn{1}{l||}{Xception~\cite{chollet2017xception}} & \multicolumn{1}{c|}{ 0.999347} & \multicolumn{1}{c|}{0.950000} & \multicolumn{1}{c}{0.996094}\\
\multicolumn{1}{l||}{MCDN~\cite{fu2018multi}} & \multicolumn{1}{c|}{0.999604} & \multicolumn{1}{c|}{0.959375} & \multicolumn{1}{c}{ 0.998438}\\
\multicolumn{1}{l||}{MSCNN~\cite{hao2019anterior}} & \multicolumn{1}{c|}{0.999727} & \multicolumn{1}{c|}{0.978125} & \multicolumn{1}{c}{0.996875}\\
\hline
\multicolumn{1}{l||}{SMA-Net} & \multicolumn{1}{c|}{\textbf{1.000000}} & \multicolumn{1}{c|}{\textbf{1.000000}} & \multicolumn{1}{c}{\textbf{1.000000}}\\
\hline
\end{tabular}}
\label{table3}
\end{table}

\subsection{Performance of open angle and angle-closure classification}

We also evaluated our method  for binary ACA classification, i.e., open angle and angle-closure cases. The public AS-OCT dataset,  Angle-closure Glaucoma Evaluation (AGE) dataset～\cite{AGE2020} was used, which includes 3200 AS-OCT images with resolution of $2130\times998$ pixels. These images were divided into two sets: 1600 training and 1600 testing. 
%We divided the training set into four subsets with different train set and test set in terms of patients. And we use the 4-fold cross validation method to train the images and get the average value of the test set on each subset.

%\vspace{0.2cm}
%\noindent\textbf{Evaluation Methods:} The evaluation metrics followed AGE Challenge are utilized to measure the binary classification performance: sensitivity $({\textit{{Sen}}})$, specificity $({\textit{{Spe}}})$. Moreover, we additionally report area under ROC curve (AUC). 

In this experiment, we changed the output of our method to binary classification.
We carried out a comprehensive comparison between the proposed and the state-of-the-art methods: Multi-Context Deep Network (MCDN)~\cite{fu2018multi}, Multi-Scale Regions Convolutional Neural Networks (MSCNN)~\cite{hao2019anterior}; Resnet-34~\cite{he2016deep}; Inception-V3~\cite{xia2017inception}; and Xception~\cite{chollet2017xception}. Table \ref{table3} shows that our SMA-Net outperforms all competing methods in terms of all metrics (AUC, \textit{Sen} and \textit{Spe}). To be more specific, it can be seen that our SMA-Net outperforms the Xception (our backbone network) alone, with improvement of $5.00\%$  in \textit{Sen}. 
Interestingly, all the methods achieved remarkable AUC scores. This is because, as we suggested  above, the binary classification of open or angle-closure is relatively easy, and this finding is also evidenced by the leader board of AGE challenge - the task of open angle and angle-closure classification by means of AS-OCT has attained remarkably high standards of performance by using state-of-the-art deep networks, with AUC scores higher than 0.98 across the board. Another reason is that AGE dataset was motivated for the classification of explicit open and closed ACA.

%It confirms that the MSDA block of our MA-Net can effectively improve performance for glaucoma classification. 

%%%%%%%%%%%%%%%%%%%%%%%%%%%%%%%%%%%%%%%%%%%%%%%%%%%%%%%%%%%%%%%%%%%%%%%%%%%%%%%
\section{Conclusion}    %  and Future Work
Most of the existing automated methods are only able to classify the ACA as either open or angle-closure. In this paper, we proposed an automated ACA classification framework, which is not only able to classify open angle and angle-closure, but is also capable of grading the three-class open-narrow-synechiae  ACA from AS-OCT imagery, so as to further guide clinical management at different stages of glaucoma.
To be more specific, we introduced a novel block, named the MSDA block, with a view to learning multi-scale discriminative representations over AS-OCT volumes. In addition, a new multi-loss function is used to combine the  slice-based and  sequence-based losses, as thus to extract spatial and temporal features from AS-OCT image sequences. The results demonstrate that the proposed method outperforms other state-of-the-art 2D and 3D deep networks. 
It may be expected that the proposed model could be a powerful tool for diagnosing the presence, and analyzing the progression of angle-closure glaucoma.

\bibliographystyle{splncs}
% \vspace{-0.3cm}
\bibliography{refs}

\end{document}